\documentclass{article}

\usepackage{microtype}
\usepackage{graphicx}
\usepackage{booktabs} % for professional tables 
\usepackage{amsmath}
\usepackage{amsfonts}
\usepackage{svg}
\usepackage{rotating}
\usepackage{multirow}
\usepackage{xspace}
\usepackage{caption}
\usepackage{subcaption}

\usepackage{hyperref}

\DeclareGraphicsExtensions{.png,.pdf}
\usepackage[arxiv]{mlsys2025}

\mlsystitlerunning{\sysname: A Practical Sparse Attention Method for Long-Context LLM Inference}

\newcommand{\sysname}{Kascade\xspace}
\newcommand{\llama}{Llama-3.1-8b-Instruct\xspace}
\newcommand{\distllama}{DeepSeek-R1-Distill-Llama-8b\xspace}
\newcommand{\qwen}{Qwen3-8b\xspace}
\begin{document}

\twocolumn[
\mlsystitle{\sysname: A Practical Sparse Attention Method for Long-Context LLM Inference}

% It is OKAY to include author information, even for blind
% submissions: the style file will automatically remove it for you
% unless you've provided the [accepted] option to the mlsys2025
% package.

% List of affiliations: The first argument should be a (short)
% identifier you will use later to specify author affiliations
% Academic affiliations should list Department, University, City, Region, Country
% Industry affiliations should list Company, City, Region, Country

% You can specify symbols, otherwise they are numbered in order.
% Ideally, you should not use this facility. Affiliations will be numbered
% in order of appearance and this is the preferred way.
\mlsyssetsymbol{equal}{*}

\begin{mlsysauthorlist}
\mlsysauthor{Dhruv Deshmukh}{msr}
\mlsysauthor{Saurabh Goyal}{msr}
\mlsysauthor{Nipun Kwatra}{msr}
\mlsysauthor{Ramachandran Ramjee}{msr}
\end{mlsysauthorlist}

\mlsysaffiliation{msr}{Microsoft Research India}

\mlsyscorrespondingauthor{Saurabh Goyal}{saurabhgoyal@microsoft.com}

% You may provide any keywords that you
% find helpful for describing your paper; these are used to populate
% the "keywords" metadata in the PDF but will not be shown in the document
\mlsyskeywords{Sparse Attention}

\vskip 0.3in

\begin{abstract}

Attention is the dominant source of latency during long-context LLM inference, an increasingly popular workload with reasoning models and RAG. We propose \sysname, a training-free sparse attention method that leverages known observations such as 1) post-softmax attention is intrinsically sparse, and 2) the identity of high-weight keys is stable across nearby layers. \sysname computes exact Top-$k$ indices in a small set of \emph{anchor} layers, then reuses those indices in intermediate \emph{reuse} layers. The \emph{anchor} layers are selected algorithmically, via a dynamic-programming objective that maximizes cross-layer similarity over a development set, allowing easy deployment across models. The method incorporates efficient implementation constraints (e.g. tile-level operations), across both prefill and decode attention. The Top-$k$ selection and reuse in \sysname is \textit{head}-aware and we show in our experiments that this is critical for high accuracy. \sysname achieves up to $4.1 \times$ speedup in decode attention and $2.2 \times$ speedup in prefill attention over FlashAttention-3 baseline on H100 GPUs while closely matching dense attention accuracy on long-context benchmarks such as LongBench and AIME-24. The source code of \sysname will be available at \url{https://github.com/microsoft/kascade}.

\end{abstract}
]

% this must go after the closing bracket ] following \twocolumn[ ...

% This command actually creates the footnote in the first column
% listing the affiliations and the copyright notice.
% The command takes one argument, which is text to display at the start of the footnote.
% The \mlsysEqualContribution command is standard text for equal contribution.
% Remove it (just {}) if you do not need this facility.

\printAffiliationsAndNotice{}  % leave blank if no need to mention equal contribution
%\printAffiliationsAndNotice{\mlsysEqualContribution} % otherwise use the standard text.

\section{Introduction}

Large language models are increasingly deployed in settings that demand long contexts: chain-of-thought style reasoning, multi-step tool use, retrieval-augmented generation over multi-document corpora, coding agents, etc. In long context inference, the computation cost is dominated by the attention operation in both the prefill (where attention is $\mathcal{O}(n^2)$ for context length $n$, compared to $\mathcal{O}(n)$ MLP operation) and decode ($\mathcal{O}(n)$ attention vs $\mathcal{O}(1)$ MLP) phases. Moreover, decode attention is memory bandwidth bound and therefore does not benefit much from batching, making it inefficient on modern GPUs.

The attention operation is expensive because each token has to attend to all previous context tokens. A common method to decrease the cost is sparse attention, where the attention function is approximated by using only a subset of the context tokens. Numerous sparse attention methods have been proposed,
including fixed-pattern~\cite{beltagy2020longformer,xiao2023efficient,zaheer2020big, jiang2024minference}, workload-aware~\cite{gim2024prompt, yao2025cacheblend, lu2024turborag, ma2025blockattention}, and dynamic sparsity variants~\cite{singhania2024loki,zhang2023h2o,tang2024quest, yang2025lserve, gao2024seerattention, gao2025seerattention}. However, some of these methods require model retraining, or sacrifice generality across tasks. 

In this paper, we present \sysname, a dynamic sparsity based technique which reduces the cost of attention significantly while retaining the accuracy of dense attention. Compared to other training free sparse attention schemes, we find that \sysname achieves the best accuracy on AIME-24, at a given sparsity ratio, as shown in Table~\ref{table:aime24}.

\sysname leverages two known observations: 1) the post-softmax attention scores are inherently sparse, and 2) the sparsity structure is stable across nearby layers. Figure~\ref{fig:oracle_attn_sparsity} shows the sparsity inherent in attention operation. As shown, only 256 (about 10\%) of the tokens contribute to over 95\% of the softmax output. This is intuitive, as the softmax operation exponentially amplifies the relative magnitude of larger values compared to the smaller ones. Thus, if we have an oracle that determines the Top-$k$ tokens, which contribute most to the attention operation, we can get a very accurate approximation of the operation, at a fraction of the cost. Figure~\ref{fig:oracle_results} shows the accuracy of Oracle Top-$k$ with varying values of $k$. As shown, with just 2.5\% of tokens, one can recover almost the full accuracy of dense attention.

However, computing these Top-$k$ values efficiently is a fundamental challenge as accurate computation will entail reading all the $\mathcal{O}(n)$ \textit{keys} of the tokens and computing the softmax. This is where we leverage the second observation --- the exact Top-$k$ of $layer_i$ is very close to the exact Top-$k$ of $layer_{i+m}$ for reasonable values of $m$. Figure~\ref{fig:cross_layer_similarity} illustrates this observation. For example, the Top-$k$ of layer 16 captures 99\% of the Top-$k$ attention of layers 17 and 18.

These observations motivate our solution:  we compute full attention, and identify Top-$k$ tokens, only on a subset of layers, which we call \emph{anchor} layers, and reuse those Top-$k$ tokens to compute sparse attention in intermediate layers. In order to identify the best subset of \emph{anchor} layers, we propose an automated dynamic programming scheme that maximizes cross layer similarity scores. This makes it easy to deploy \sysname on new models. We also observe that the Top-$k$ tokens vary across heads, and propose a head remapping technique when reusing these Top-$k$ tokens. We find this is critical for high accuracy. Therefore, by algorithmically choosing a good set of \emph{anchor} layers, and being head-aware, \sysname achieves the best accuracy on AIME-24 among similar techniques for a given sparsity ratio.

\sysname incorporates design decisions based on low-level kernel implementation of the attention operation to get strong performance gains in both prefill and decode. For example, in the prefill attention kernel, the $QK^{T}$ operation is performed over tiles for maximizing parallelization and efficiency. As a result, consecutive tokens of a sequence in a \textit{Q}-tile, share the same \textit{key} tokens. Thus, \sysname performs the Top-$k$ selection at a tile level and not independently for each token. \sysname kernels are implemented in TileLang~\cite{wang2025tilelang} and deliver substantial efficiency gains on H100s (up to $4.1 \times$ for decode attention compared to FlashAttention-3 baseline) with negligible impact on task accuracy across models and benchmarks.

In summary, we make the following contributions:
\begin{itemize}
\item \sysname{} introduces three efficient mechanisms: tiled top-K, head remapping and automatic anchor layer selection, for making sparse attention practical and accurate.

\item Compared to previous techniques, \sysname{} achieves higher accuracy at the same Top-$k$. For example, in AIME-24, \sysname delivers substantially higher accuracy ($8\textbf{--}10$\% absolute) compared to previous schemes with two different models at 10\% Top-$k$.

\item By automating anchor layer selection and head remapping, and with a performant kernel for H100s, we believe \sysname{} is the first kernel that makes it easy to deploy sparse attention for different models.

\item On H100s, \sysname{} delivers up to $4.1\times$ faster performance than FlashAttention-3 decode kernel and up to $2.1\times$ faster performance than FlashAttention-3 prefill kernel, thereby ensuring significant performance gains while achieving comparable accuracy in long-context tasks.

\end{itemize}
\section{Background and Related Work}
\label{sec:background}

\subsection{Scaled Dot-Product Attention}
\label{subsec:background_attention}

In scaled dot-product attention~\cite{vaswani2017attention}, each new token must attend to all previous tokens. Formally, for a query $\boldsymbol{q}_t \in \mathbb{R}^{d}$, corresponding to the current token, and key-value pairs $\boldsymbol{K}, \boldsymbol{V} \in \mathbb{R}^{N \times d}$ of past tokens, the attention output $\boldsymbol{y}$ is computed as:
\begin{align}
\label{eq:attn_p}   \boldsymbol{p} &= \mathrm{softmax}\Bigl(\frac{\boldsymbol{q}_t \cdot \boldsymbol{K}^\top}{\sqrt{d}}\Bigr) \in \mathbb{R}^{N} \\
\label{eq:attn_o}   \boldsymbol{y} &= \boldsymbol{p} \cdot \boldsymbol{V} \in \mathbb{R}^{d}
\end{align}

Here $N$ is the current sequence length, and $d$ is the hidden dimension. Equation~\ref{eq:attn_p} computes a weight for every token, and Equation~\ref{eq:attn_o} computes the output as a weighted sum of values in $\boldsymbol{V}$. As we'll see in Section~\ref{subsec:method_oracle_topk}, the $\boldsymbol{p}$ vector is sparse, and most of the weights are close to 0.

The above operation results in $O(N)$ computation and memory access per token, making attention the dominant contributor to latency in long-context inference. 

\subsection{Grouped Query Attention}

In Multi-Head Attention~\cite{vaswani2017attention}, each head performs independent attention with its own learned projections of $Q, K, V$, and their outputs are concatenated and linearly transformed. Grouped Query Attention (GQA)~\cite{ainslie2023gqa} generalizes this formulation by allowing multiple query heads to share a common set of key-value projections, reducing memory bandwidth and improving efficiency in both training and inference. GQA is widely adopted in recent models, but imposes additional constraints on efficient implementation of sparse attention schemes, which we explore in Section~\ref{subsec:method_query_pooling}.

\subsection{Sparse Attention Methods}

As we observed in Section~\ref{subsec:background_attention}, the attention operation is expensive because each token has to attend to all previous context tokens. To address this bottleneck, various sparse attention mechanisms have been proposed.

\emph{Fixed pattern sparsity} approaches fix a connectivity pattern (e.g., sliding windows plus a small set of global tokens)~\cite{beltagy2020longformer,xiao2023efficient,zaheer2020big, jiang2024minference} and compute attention only over these tokens. Some models deploy sliding window attention on a subset of layers to limit the attention cost~\cite{team2024gemma,agarwal2025gpt}. However, these approaches work best when baked into the architecture before pre-training, as the model needs to learn to attend within this connectivity pattern; or require some amount of post-training.

Another class of techniques is \emph{workload-aware sparsity} which primarily targets RAG like scenarios and limits the attention operation of document tokens to within the context of the same document~\cite{gim2024prompt, yao2025cacheblend, lu2024turborag, ma2025blockattention}. These techniques may also require some post-training to maintain accuracy, and they only optimize the prefill phase of inference.

The final family of approaches is \emph{dynamic sparsity}, where a subset of \textit{k} tokens is dynamically selected for the attention computation~\cite{ribar2023sparq, singhania2024loki,zhang2023h2o,tang2024quest, yang2025lserve, gao2024seerattention, gao2025seerattention}. Selecting the best tokens efficiently is, however, an open research problem.

\subsection{Cross-Layer Similarity in Attention}

One key observation we use, described in Section~\ref{subsec:method_cls}, is that the sparsity patterns of attention weights exhibit strong correlations across neighboring layers. Prior works \cite{ying2021lazyformer, bhojanapalli2021leveraging, xiao2019sharing} have evaluated sharing full attention weights across layers, which we found to degrade accuracy.

Some recent works like OmniKV~\cite{hao2025omnikv}, TidaDecode~\cite{yang2024tidaldecodefastaccuratellm}, and LessIsMore~\cite{yang2025moretrainingfreesparseattention} have also used this observation to design an approximate Top-$k$ attention scheme. OmniKV has a focus on reducing memory capacity requirements, and hence offloads the KV cache to the CPU. The benefit in memory capacity, however, comes at the cost of reduced performance because of CPU-GPU transfers. LessIsMore, built on top of TidalDecode, computes Top-$k$ indices on fewer layers chosen manually. A key challenge with these schemes is that there is not automated way to identify the \emph{anchor} layers which makes it difficult to deploy to new models. These schemes use a shared set of Top-$k$ indices across all heads, while we find separate Top-$k$ indices for every key head to improve accuracy, as described in Section~\ref{subsec:method_head_remapping}.
\section{\sysname}

We now outline the various insights and techniques needed for a complete design and implementation of \sysname.

\subsection{Oracle Top-$k$ Selection}
\label{subsec:method_oracle_topk}

We first ask a feasibility question: can we approximate attention using only a small subset of past tokens without losing task accuracy? Since the softmax operation in Equation~\ref{eq:attn_p} exponentially amplifies the relative magnitude of larger values compared to the smaller ones, it has an inherent sparsification. To exploit this sparsification, we define an \textit{Oracle} Top-$k$, where we compute the attention, in Equation~\ref{eq:attn_o}, only over $k$ tokens with the highest $\boldsymbol{p}$ values. Since computing these $k$ tokens requires the full softmax operation, we use the term \emph{Oracle}, and this provides an accuracy upper bound.

\begin{figure}[t!]
    \centering
    \includegraphics[width=\linewidth]{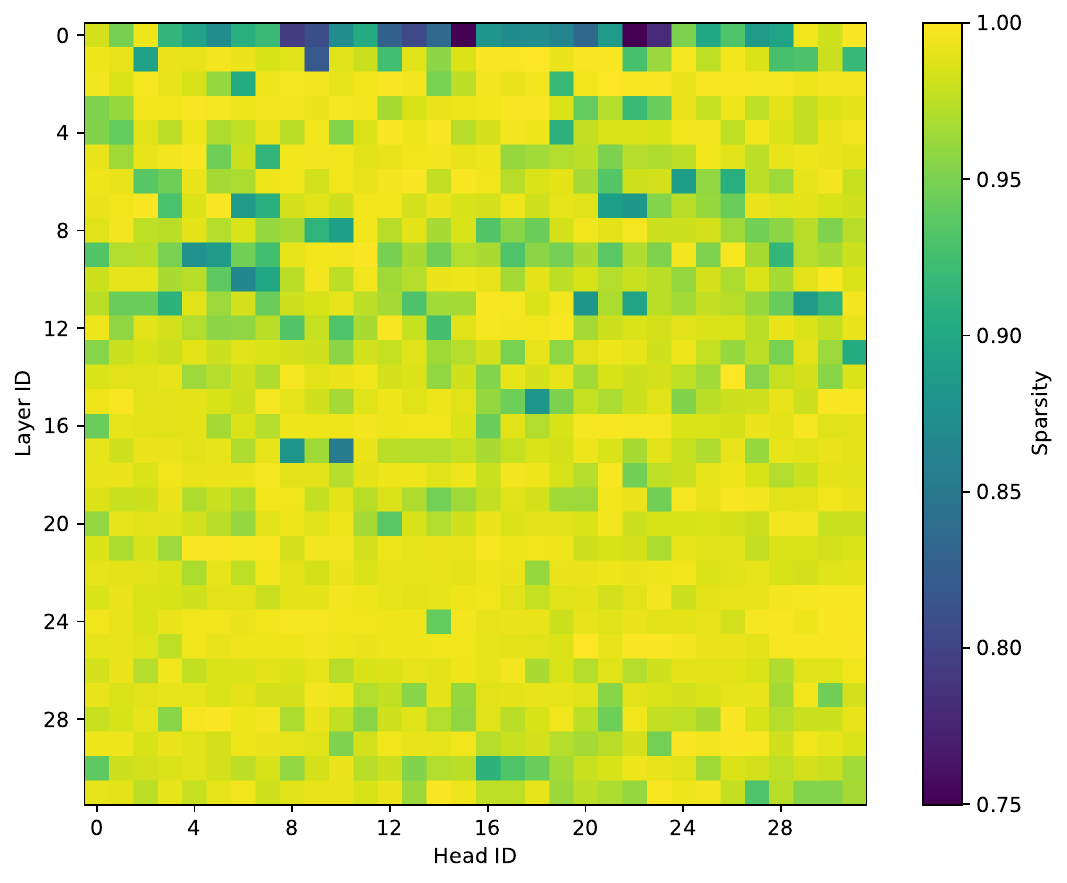}
    \caption{Attention weight covered by top 256 keys across layers and heads. Except for layer 0 rest of layers have high sparsity across majority of the heads. Model=\llama, Dataset=MuSiQue.}
    \label{fig:oracle_attn_sparsity}
\end{figure}

Figure~\ref{fig:oracle_attn_sparsity} confirms that the output of softmax is indeed sparse. For example, 95\% of the total attention mass in almost all layers and heads is captured by the top 256 tokens. The only exception is layer 0, where the distribution is considerably flatter. \sysname therefore always computes full dense attention in layer 0 and applies sparsity only layer 1 onwards. 

\begin{figure}[ht]
    \centering
    \includegraphics[width=\linewidth]{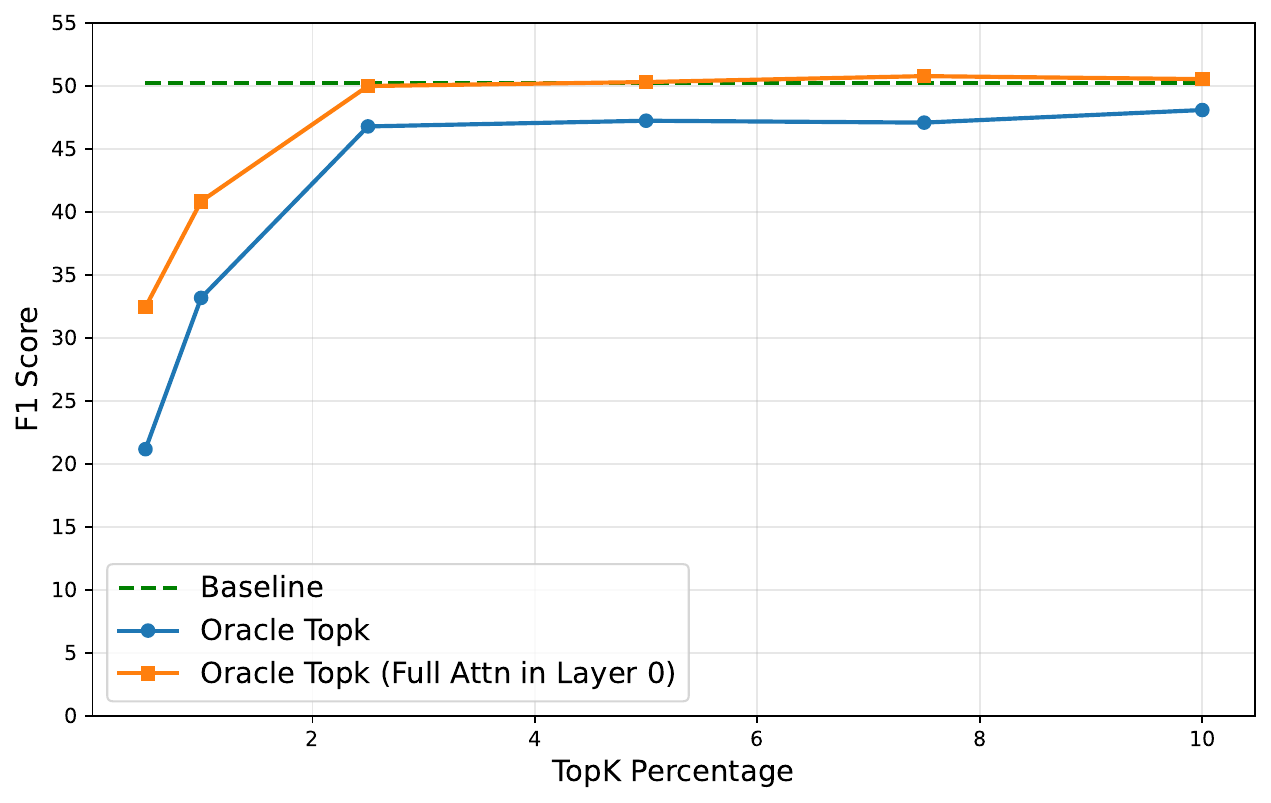}
    \caption{Oracle Top-$k$ attention results with varying Top-$k$ percentage. With layer 0 doing full attention, Oracle Top-$k$ matches baseline score even with Top-$k$ as 5\%.  Model=\llama, Dataset=2WikiMultihopQA.}
    \label{fig:oracle_results}
\end{figure}

Figure~\ref{fig:oracle_results} evaluates how aggressively we can sparsify while preserving task quality. We replace dense attention with Oracle Top-$k$ attention, and measure end-task accuracy (F1 on 2WikiMultihopQA\cite{xanh2020_2wikimultihop} for \llama). Even at $k/N=2.5\%$, Oracle Top-$k$ attention matches the accuracy of full attention. This shows that, if Top-$k$ can be estimated efficiently, the performance upside can be significant without compromising accuracy.

\sysname is designed to approximate this oracle efficiently. The next subsections address two challenges of doing this at runtime: (1) computing the Top-$k$ set without first materializing full attention, and (2) enforcing GPU-friendly sharing of Top-$k$ indices across tiles, layers, and heads for high throughput. We address (1) using cross-layer reuse of Top-$k$ indices and (2) using head remapping, tile-level pooling, and GQA-aware constraints.

\subsection{Cross-Layer Similarity}
\label{subsec:method_cls}
%To motivate \sysname, we analyze the overlap of Top-$k$ key selections between layers.
We now ask whether we can avoid recomputing the Top-$k$ set independently in every layer by reusing it across nearby layers. For a query token $q$, let $P^{(l, h)}_q \in \mathbb{R}^N$ denote the post-softmax attention distribution, at layer $l$ and head $h$, where $N$ is the current sequence length. We define the layer attention distribution $P^l_q$ as the average of $P^{(l, h)}_q$ across all heads. We then define the Top-$k$ index set for token $q$ at layer $l$ as $I^l_q = topk(P^l_q, k)$.

To quantify how well the Top-$k$ set from one layer can be reused in another layer, we define a similarity score between two layers $a, b$ where $a < b$. For token $q$, we measure how much of layer $b$'s oracle attention mass would be recovered if we were to force layer $b$ to use the Top-$k$ keys selected at layer $a$:

\begin{equation}
sim(a,b)_q = \frac{\sum_{i=1}^{k} P^b_q[I^a_q[i]]}{\sum_{i=1}^{k} P^b_q[I^b_q[i]]},
\label{eq:sim_matrix}
\end{equation}
\[
\text{where $a > b$ and $|I^a_q| = |I^b_q| = k$}
\]
Values near 1 indicate that the identity of the high-importance keys is stable across layers. We compute this similarity for each query token in a prompt, then average across all tokens in that prompt. We then average again across a development set of multiple prompts to obtain a layer-by-layer similarity matrix $S \in \mathbb{R}^{L \times L}$, where $L$ is the total number of layers.

\begin{figure}[ht]
    \centering
    \includegraphics[width=\linewidth]{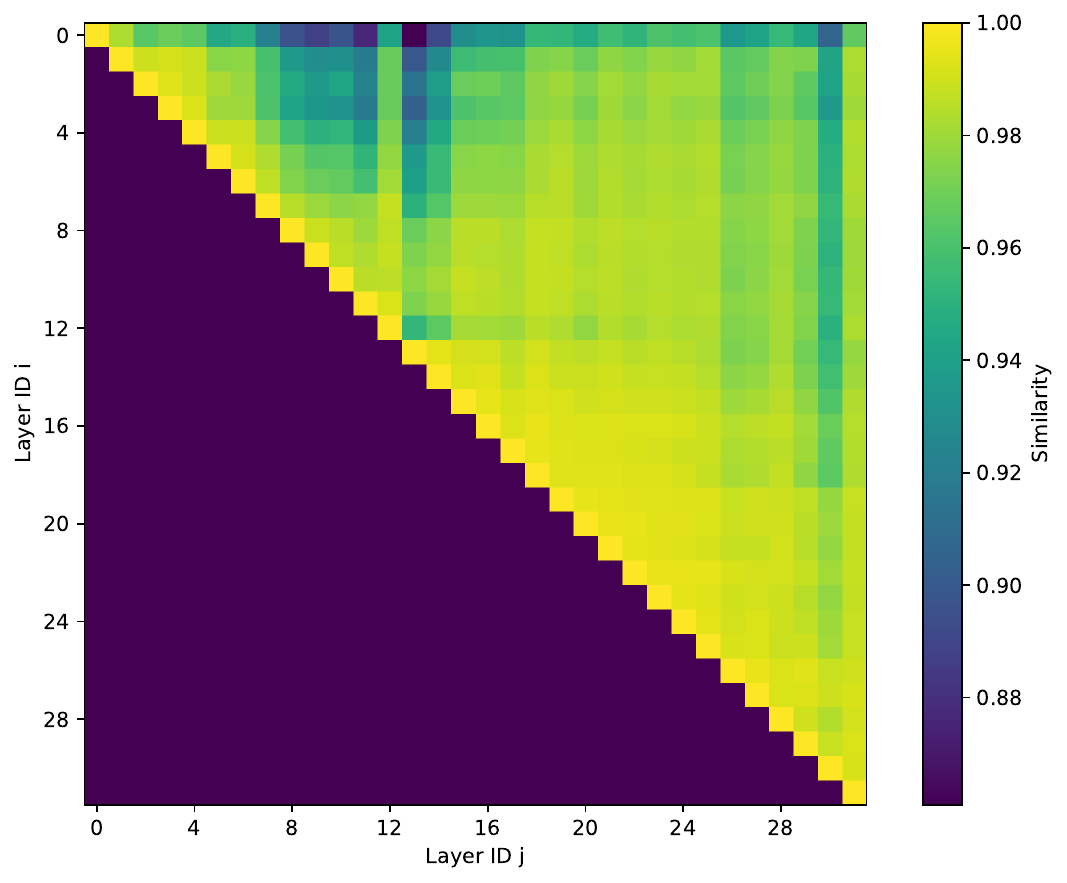}
    \caption{Cross layer similarity using top 256 keys. Bright cell indicates that Top-$k$ indices of layer i cover high fraction of attention covered by Top-$k$ indices of layer j itself. Model=\llama, Dataset=MuSiQue}
    \label{fig:cross_layer_similarity}
\end{figure}

Figure~\ref{fig:cross_layer_similarity} shows this cross-layer similarity matrix for \llama using MuSiQue as the development set, with $k=256$ (the average context length in MuSiQue is 2.3K). Most adjacent layer pairs achieve similarity scores close to 1. Similarity generally decays with layer distance, but remains high across short ranges. For example, similarity score of most nearby pairs stays above 0.98, meaning that more than 98\% of the oracle Top-$k$ attention mass at layer $b$ is already covered by the Top-$k$ keys chosen at layer $a$.  

With this observation, \sysname computes Top-$k$ indices on only a small set of \emph{anchor layers}. These indices are then used to compute Top-$k$ attention on for the next few \emph{reuse} layers.

\subsection{Anchor Layer Selection}
\label{subsec:method_anchor_layer_selection}

Given a budget for the number of anchor layers, we want to select the set of anchor layers, such that it maximizes the similarity between the chosen anchor layers and the corresponding reuse layers. \sysname performs this selection using the dynamic programming algorithm, shown in Algorithm~\ref{alg:anchor_dp} which uses the similarity matrix $S$ as input. To construct $S$, we evaluate the similarity scores $\text{sim}(a,b)_q$ (equation~\ref{eq:sim_matrix}) for every token $q$ in a prompt, then take the \emph{minimum} across tokens in that prompt, rather than the mean. This makes the score conservative and ensures that the similarity is determined by the worst token in a prompt. We observed that this resulted in a more robust anchor selection. We used $k=64$ for computing the similarity scores and found it to work well across experiments. The similarity matrix also incorporates the modifications described in Section~\ref{subsec:method_query_pooling}, and Section~\ref{subsec:method_head_remapping}.

\begin{algorithm}[ht]
  \caption{Anchor Layer Selection}
  \label{alg:anchor_dp}
  \begin{algorithmic}[1]
    \REQUIRE Similarity matrix $S$, budget $M$, layers $1\ldots L$
    \STATE Initialize $\mathrm{dp}[][] = -\infty, \mathrm{path}[][]=0$
    \STATE $\mathrm{dp}[1][1] = S[1][1]$
    \FOR{$m=2 \to M+1$}
      \FOR{$j=m \to L+1$}
        \STATE $\mathrm{dp}[m][j] = \max_{i=m-1}^{j-1}(\mathrm{dp}[m-1][i] + \sum_{l=i}^{j-1} S[i][l])$
        \STATE $\mathrm{path}[m][j] = argmax(.)$
      \ENDFOR
    \ENDFOR
    \STATE Backtrack on $\mathrm{path}[M+1][L+1]$ to recover $\{\ell\}$
  \end{algorithmic}
\end{algorithm}

\begin{figure}[ht]
    \centering
    \includegraphics[width=\linewidth]{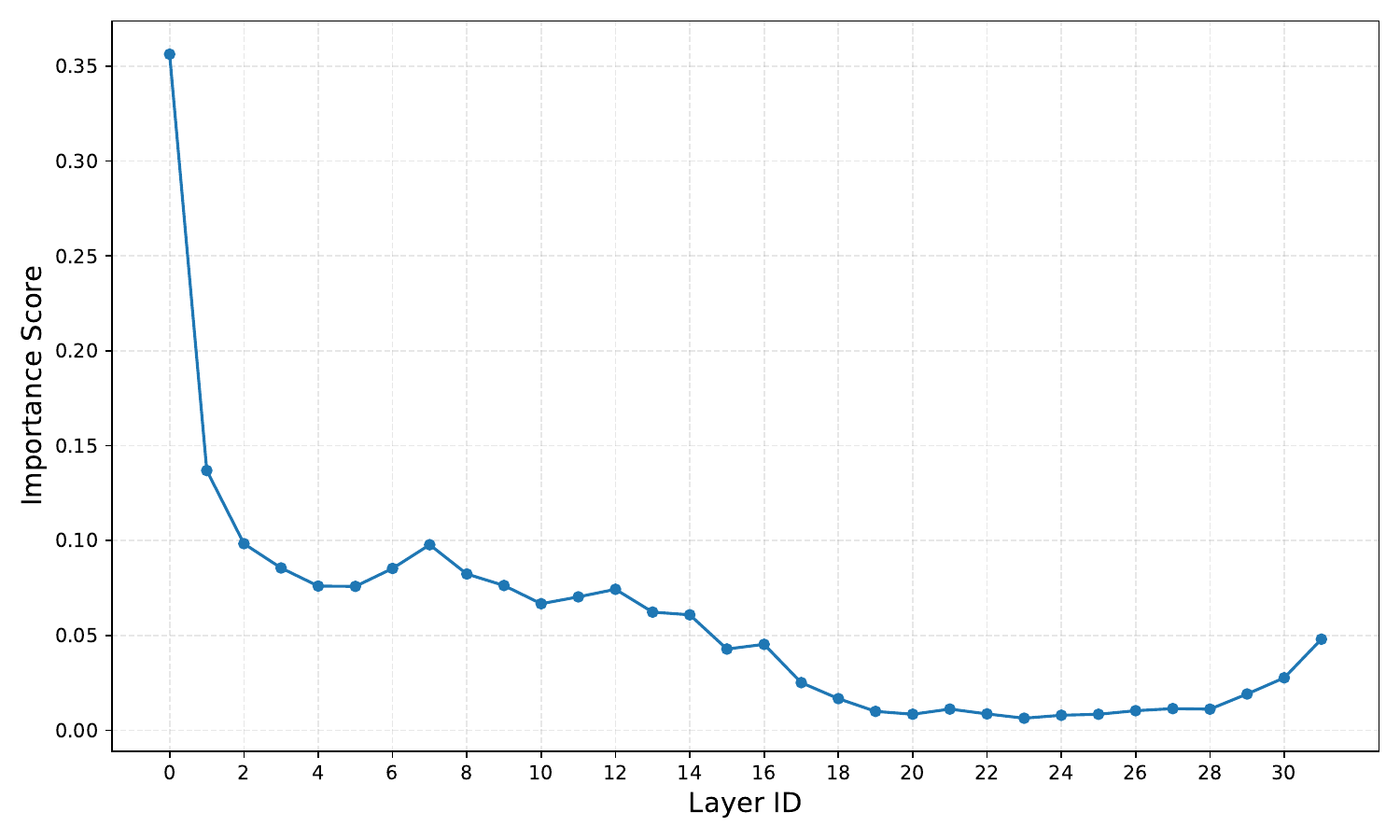}
    \caption{Importance scores of attention blocks of all layers. Deeper layers have lower importance than the initial layers. Layer 0 has highest importance. Model=\llama, Dataset=MuSiQue.}
    \label{fig:layer_attn_importance}
\end{figure}

Prior work has observed that attention in deeper layers can be less important than attention in earlier layers~\cite{he2024matters}. We account for this observation by assigning each layer $l$ an importance weight $w_l$. If $\boldsymbol{x}^i_l$ and $\boldsymbol{y}^i_l$ are an (input, output) pair of attention at layer $l$, we define the importance score $w_l^i$ as:
\[
w_l^i = 1 - CosineSim(\boldsymbol{x}^i_l, \boldsymbol{y}^i_l)
\]
Intuitively, if attention barely changes the representation (high cosine similarity), that layer’s attention block matters less. The layer's weight is computed by aggregating this score over the same development set. The similarity matrix is then weighted by this importance measure.
\[
sim[i][j] = w_j \cdot sim[i][j]
\]
Figure~\ref{fig:layer_attn_importance} shows the importance score for all layers in the \llama model showing a sharp decrease in importance of deeper layers.

We now discuss more enhancements to \sysname to incorporate finer attention implementation details.

\subsection{Query pooling}
\label{subsec:method_query_pooling}

Most modern LLMs use GQA~\cite{ainslie2023gqa} where multiple query heads share the same KV-heads. For efficiency and maximizing parallelization, the decode GQA attention kernels construct Q-tiles (for the $QK^T$ computation) by combining query values of all query heads sharing the KV-heads. Similarly, prefill kernels construct Q-tiles by combining consecutive tokens of the prompt as they all share the same prefix for the $QK^T$ operation. This allows batching of memory loads and reusing the fetched $K$ values across multiple queries, resulting in higher GPU utilization.

\begin{figure}[t!]
    \centering
    \includegraphics[width=\linewidth]{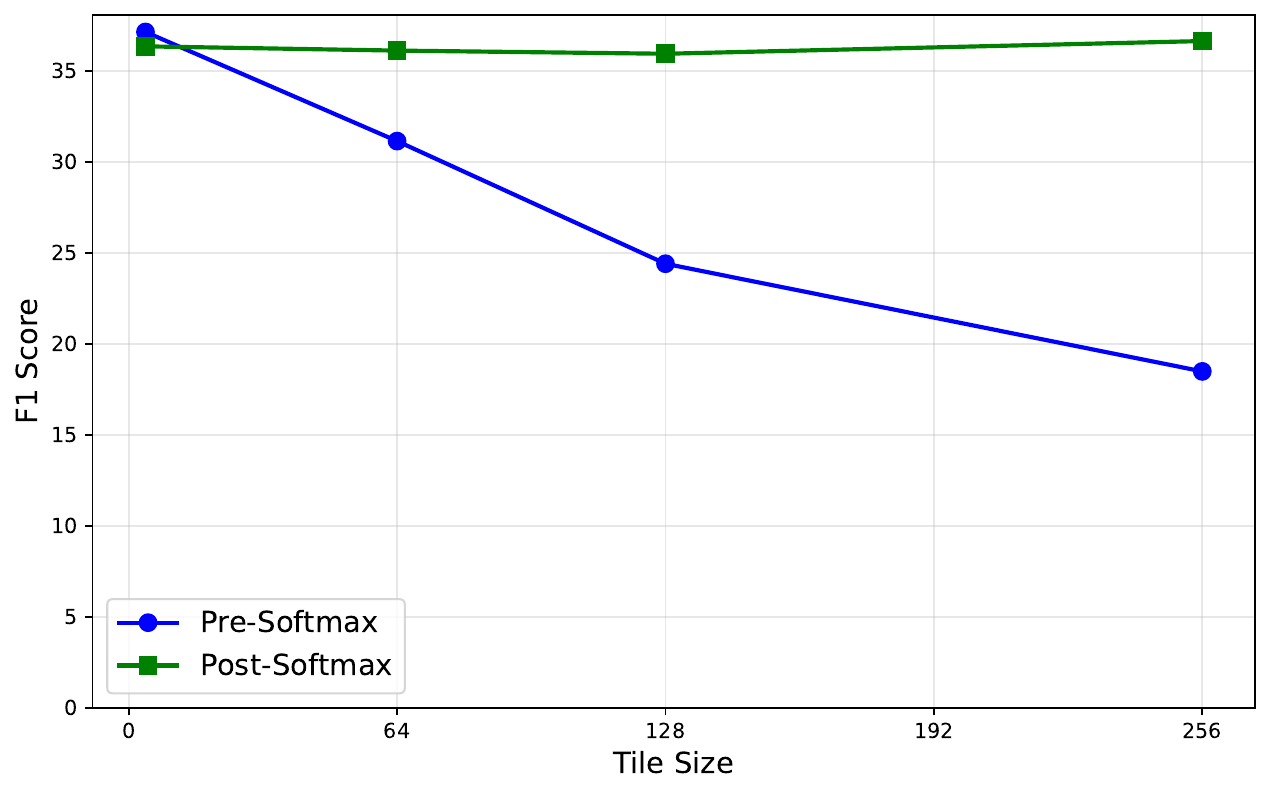}
    \caption{Comparison of Top-$k$ attention accuracy, when pooling with Pre vs Post Softmax attention scores, across different tile sizes. Top-$k$ percentage here is $10\%$. The smallest tile size is $4$ where only the queries corresponding to the same key head are pooled. Post Softmax is more robust to changes in tile size and does consistently well across all tile sizes. Model=\llama, Dataset=MuSiQue}
    \label{fig:oracle_with_pooling_results}
\end{figure}

In order to maintain this efficiency, \sysname needs to ensure that all query tokens in a tile share the same Top-$k$ indices. We must thus construct a single ``pooled" attention score for these tiles during the Top-$k$ computation in the anchor layers. We consider two pooling strategies. \textit{Pre-Softmax} pooling constructs a pooled query representation (by averaging the query vectors in a tile), and computes attention once using this pooled query. \textit{Post-Softmax} pooling instead computes the full post-softmax attention distribution independently for each query in the tile, and then pools these attention distributions across the tile. We evaluate these pooling strategies in the Oracle setting in Figure~\ref{fig:oracle_with_pooling_results}. As shown, Post-Softmax pooling maintains accuracy even for large tiles, while Pre-Softmax pooling degrades as tile size increases.

Based on these results, \sysname adopts Post-Softmax pooling. In decode, we pool only across the query heads that share a key head (GQA pooling). In prefill, we pool across full tiles of 128 queries (including the GQA grouping), which matches the tile size used in our dense FlashAttention-style baselines. This choice lets \sysname reuse a single Top-$k$ index set per GQA group in decode and per tile in prefill, and therefore aligns sparse attention with the kernel structure used by high-throughput implementations.

\subsection{Head Remapping and Reuse}
\label{subsec:method_head_remapping}

\sysname computes Top-$k$ indices at the granularity of a key head. Thus, each anchor layer will have $H$ Top-$k$ sets of indices, where $H$ is the number of key heads. This raises the question --- which head's Top-$k$ index set in an \textit{anchor} layer should be mapped to a given head in a corresponding \textit{reuse} layer? One option is to perform a simple 1:1 mapping where the $i$'th heads in the \textit{anchor} and \textit{reuse} layers are mapped to each other. However, nothing in the transformer architecture requires that the head $i$ of one layer be similar to the head $i$ of another layer. We explore two strategies to handle this. One strategy is to use a shared set of Top-$k$ indices for all heads, by pooling the attention weights across all heads in a layer. This method will not account for any variation in the Top-$k$ indices across heads. In the second strategy, instead of forcing a single shared Top-$k$ across all heads, we build an explicit mapping from heads in each \textit{reuse} layer to the most similar head in the corresponding \textit{anchor} layer. For computing this mapping, we use the same similarity score defined earlier, but at a head level, and find a head mapping that maximizes the similarity. Note that this mapping can be a many-to-one mapping. A similar technique is proposed in \cite{bhojanapalli2021leveraging} for reusing complete attention weights across layers.

\begin{figure}[t!]
    \centering
    \includegraphics[width=\linewidth]{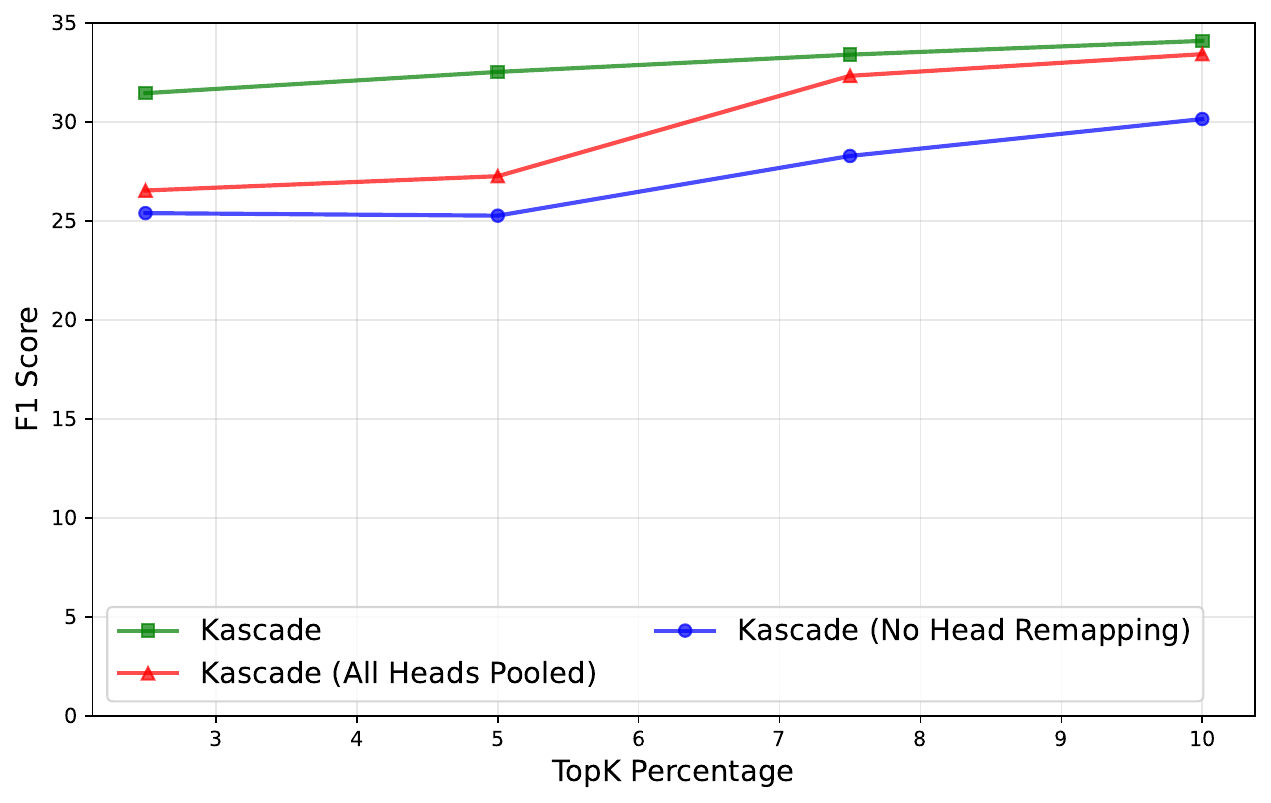}
    \caption{Comparison of \sysname variants with head remapping, without head remapping and pooling across all heads, for different Top-$k$ percentages. The tile size is $128$, which is our default tile size for prefill. No remapping is the worst. Head remapping gives consistent scores across all Top-$k$ percentages thus providing a larger operating range than pooling across all heads. Model=\llama, Dataset=MuSiQue}
    \label{fig:head_mapping_results}
\end{figure}

Figure~\ref{fig:head_mapping_results} compares these two strategies for \llama on MuSiQue across a range of Top-$k$ budgets. We observe that the head-remapping approach is consistently more robust, especially at smaller values of Top-$k$. \sysname therefore uses head remapping by default, but we also report results for the variant with a shared Top-$k$, across all heads, in Section~\ref{subsec:eval_accuracy} for completeness.

\subsection{Efficient Kernel Implementation}
\label{subsec:method_eff}

We implement \sysname by modifying the FlashAttention \cite{shah2024flashattention} kernels for prefill and decode, in TileLang \cite{wang2025tilelang}. TileLang is a tile-level programming language for GPU kernels, which also takes care of various optimizations like pipelining, memory layouts and specialized instructions. We use both the original FA3, as well as TileLang FlashAttention kernels as our baseline in performance evaluations.

For intermediate \emph{reuse} layers, we pass the Top-$k$ indices from the previous \emph{anchor} layer, and the head mapping computed on the development set. During attention computation, we load the keys by consulting the Top-$k$ indices. The key loads that make a key tile are not contiguous, but given that each key is large, about 256 bytes, we do not notice any overhead with this. This is in contrast to claims by block sparse attention approaches \cite{tang2024quest, gao2024seerattention}.

For \emph{anchor} layers, we use a multi pass approach. Since we use Post-Softmax pooling, we need to first compute the post-softmax attention weights for each $q$ separately, and then pool them across a tile. We can not compute the pooled attention weights in one pass, since softmax operation requires knowing the full row sum.
\begin{itemize}
    \item The first pass computes the full attention weight matrix $QK^T$, as well as the row sum vector ($\sum_{j=1}^{m} QK^T_{ij}$). This does about half the work of full attention. In decodes, we write out both these to HBM. In prefill, since the attention weight matrix is large, we only output the row sum vector.
    \item The second pass outputs the pooled post-softmax attention weights for each $Q$-tile. For decodes, we read the weights from the first pass, do a softmax, and pool. For prefill, we have to recompute the attention weights, but as we know the row sum vector, we can also compute the post-softmax weights and pool them.
    \item The third pass computes the Top-$k$ indices over the output of the second pass.
    \item In the final pass, we compute Top-$k$ attention similar to \emph{reuse} layers. 
\end{itemize}
For \emph{anchor} layer 0, we need full dense attention as described in Section~\ref{subsec:method_oracle_topk}, so we compute dense attention in the first pass, and omit the last pass. Figure~\ref{fig:anchor_layer_time_split} shows the time split across these passes. We see that the recomputation in the second pass, for prefill, is a significant cost. We report performance numbers in Section~\ref{subsec:eval_perf}

\section{Evaluation}
\label{sec:eval}

\begin{table*}[ht]
\caption{Results on Longbench. For StreamingLLM, sliding window is set to 30\% with 4 sink tokens. For the Top-$k$ attention methods Top-$k$ is set to 10\%.
Note that for Quest, OmniKV, and LessIsMore, the prefill phase uses full attention as they only optimize the decode. Since longbench is prefill-heavy, the high accuracy obtained by these schemes is not unexpected while \sysname{} achieves high accuracy while optimizing both the prefill and decode for this benchmark.}
\label{table:longbench}
\begin{center}
\resizebox{\linewidth}{!}{%
\begin{tabular}{c|l|cccccc|c} 
\toprule
\multicolumn{1}{l|}{Model}                                                                                                         & Strategy                      & SQA  & MQA   & Summ.  & Fewshot        & Synthetic      & Code           & \multicolumn{1}{c}{Avg.}  \\ 
\midrule
\multirow{9}{*}{\begin{sideways}\begin{tabular*}{4.5\normalbaselineskip}{c}Meta-Llama-3.1\\-8B-Instruct\end{tabular*}\end{sideways}} 
& Baseline (Dense) & 48.43 & 43.18 & 25.99 & 63.22 & 34.83 & 59.89 & 45.92 \\
 & StreamingLLM & 24.83 & 25.05 & 22.41 & 56.33 & 12.00 & 59.89 & 33.42 \\
 & LessIsMore (decode-only) & 48.15 & 42.71 & 25.38 & 63.05 & 34.67 & 59.16 & 45.52 \\
 & OmniKV (decode-only) & 48.22 & 43.05 & 25.97 & 63.22 & 34.72 & 59.33 & 45.75 \\
 & Quest (decode-only) & 46.97 & 42.82 & 25.71 & 62.33 & 34.14 & 54.36 & 44.39 \\
 & \textbf{\sysname} & 47.41 & 39.84 & 25.21 & 61.32 & 33.67 & 62.70 & 45.02 \\
 % & \textbf{\sysname (Decode Only)} & 48.07 & 42.93 & 25.32 & 62.76 & 34.67 & 59.17 & 45.49 \\
 & \textbf{\sysname (All Heads Pooled)} & 47.83 & 40.50 & 25.34 & 63.09 & 34.50 & 62.95 & 45.70 \\
\midrule
\multirow{8}{*}{\begin{sideways}\begin{tabular*}{3\normalbaselineskip}{c}Qwen3-8B\end{tabular*}\end{sideways}} 
& Baseline (Dense) & 47.56 & 41.35 & 24.15 & 64.32 & 34.83 & 65.90 & 46.35 \\
 & StreamingLLM & 24.33 & 28.44 & 21.01 & 56.82 & 12.50 & 64.32 & 34.57 \\
 & LessIsMore (decode-only) & 40.87 & 38.47 & 23.03 & 62.38 & 34.67 & 63.21 & 43.77 \\
 & Quest (decode-only) & 44.71 & 40.46 & 24.34 & 62.72 & 33.94 & 57.53 & 43.95 \\
 & \textbf{\sysname} & 44.19 & 40.38 & 23.02 & 60.83 & 35.00 & 63.98 & 44.57 \\
 % & \textbf{\sysname (Decode Only)} & 46.83 & 41.00 & 23.34 & 63.82 & 35.17 & 66.05 & 46.03 \\
 & \textbf{\sysname (All Heads Pooled)} & 44.87 & 42.34 & 23.74 & 61.99 & 34.50 & 62.71 & 45.02 \\
\bottomrule
\end{tabular}
}
\end{center}
\end{table*}
\begin{table}[ht]
\caption{Results on AIME-24. For StreamingLLM, sliding window is set to 30\% with 4 sink tokens. For the Top-$k$ attention methods, Top-$k$ is set to 10\%. \sysname gives the best accuracy across the board performing well even for decode heavy tasks. StreamingLLM fails to solve even a single question correctly.}
% \caption{Results on AIME-24 (decode-heavy). We report Avg.\ Pass@1 with average decode length in parentheses. Dense baseline and StreamingLLM (30\% sliding window, 4 sink tokens) are shown as references. For Top-$k$ methods, we report $k\in\{10\%,20\%\}$. At Top-$k$ of 20\%, \sysname's accuracy almost matches that of the dense baseline. StreamingLLM fails to solve even a single question correctly.}
\begin{center}
\label{table:aime24_a}
\footnotesize
\setlength{\tabcolsep}{3pt}
\begin{tabular}{@{}l | r | r@{}} 
\toprule
Strategy & \multicolumn{2}{c}{Avg. Pass@1 (Decode Length)} \\ 
\cmidrule{2-3}
& \multicolumn{1}{c|}{\begin{tabular}{@{}c@{}}DeepSeek-R1-\\Distill-Llama-8B\end{tabular}} & \multicolumn{1}{c}{Qwen3-8B} \\ 
\midrule
Baseline (Dense)   & 50.42 (11.3k) & 73.75 (14.4k) \\
StreamingLLM       & 0.00 (\hspace{0.5em}7.5k)  & 0.00 (\hspace{0.5em}6.9k)  \\
% \midrule
% \multicolumn{3}{c}{Top-$k$ = 10\%} \\
% \midrule
LessIsMore         & 36.25 (14.8k) & 60.83 (17.9k) \\
OmniKV             & 39.58 (12.5k) & -      \\
Quest              & 7.50  (22.9k) & 25.33 (28.8k) \\
\textbf{\sysname}            & \textbf{47.92} (14.6k) & \textbf{70.42} (15.9k) \\
% \textbf{\sysname (Decode Only)}    & 45.42 (14.0k) & 70 (15.7k) \\
\textbf{\sysname (All Heads Pooled)}   & 41.25 (14.0k) & 65.83 (17.9k) \\
% \midrule
% \multicolumn{3}{c}{Top-$k$ = 20\%} \\
% \midrule
% LessIsMore         & 42.92 (13.4k) & 67.83 (15.8k) \\
% OmniKV             & 46.67 (11.8k) & -      \\
% Quest              & 17.92  (17.0k) & 53.33 (23.1k) \\
% \textbf{\sysname}            & 49.17 (12.8k) & \textbf{72.92} (14.3k) \\
% % \textbf{\sysname (Decode Only)}    & 45.42 (14.0k) & 70 (15.7k) \\
% \textbf{\sysname (All Heads Pooled)}   & \textbf{50.00} (12.5k) & 65.00 (15.4k) \\
\bottomrule
\end{tabular}
\end{center}
\label{table:aime24}
\end{table}

\subsection{Setup}

We evaluate \sysname on two popular long context benchmarks, LongBench and AIME-24. We also choose two, widely used, long context models for evaluation: \llama~\cite{grattafiori2024llama}, and \qwen~\cite{yang2025qwen3}. Both these models use GQA and can support up to 128k context length. For AIME-24 evaluations, instead of \llama, we use \distllama~\cite{guo2025deepseek} which is a fine-tuned version of Llama-3.1-8b for reasoning tasks. The original \llama has very low baseline accuracy on AIME-24.

We do accuracy comparisons with dense attention, Quest~\cite{tang2024quest}, StreamingLLM~\cite{xiao2023efficient}, OmniKV~\cite{hao2025omnikv} and LessIsMore~\cite{yang2025moretrainingfreesparseattention}. We have implemented these algorithms in our own code-base, borrowing from any publicly available code. We also evaluate a variant of \sysname where we use a shared Top-$k$ across all heads (Section~\ref{subsec:method_head_remapping}.

\noindent\begin{table*}[t]
\centering
\caption{The time for \sysname is weighted average of times for \emph{anchor} layer 0, \emph{anchor} and reuse columns where the weights are $\frac{1}{32}$, $\frac{4}{32}$ and $\frac{27}{32}$. For decodes, batch size is 64, except at 512k, where it is 32. For prefills, batch size is 1. \sysname gives best speedups at Top-$k$ set to 10\%. Speedups on both original FA3 and Tilelang(TL) implementation of FA3 are shown.}\label{table:perf_full}
\scriptsize
\begin{tabular}{c|r|c|rr|rrrrrrr|rr} 
\toprule
\multicolumn{1}{l|}{Step} & Seqlen & Topk\% & FA3 & \begin{tabular}[c]{@{}c@{}}Tilelang\\ (TL)\end{tabular} & \multicolumn{2}{c}{Anchor layer 0} & \multicolumn{2}{c}{Anchor} & \multicolumn{2}{c}{Reuse} & \sysname & \multicolumn{2}{l}{Attn Speedup} \\ 
\cline{4-14}
\multicolumn{1}{l|}{} &  &  & \begin{tabular}[c]{@{}l@{}}Time\\(ms)\end{tabular} & \begin{tabular}[c]{@{}l@{}}Time\\(ms)\end{tabular} & \begin{tabular}[c]{@{}l@{}}Time\\(ms)\end{tabular} & \begin{tabular}[c]{@{}l@{}}Ratio \\w.r.t. TL\end{tabular} & \begin{tabular}[c]{@{}l@{}}Time\\(ms)\end{tabular} & \begin{tabular}[c]{@{}l@{}}Ratio \\w.r.t. TL\end{tabular} & \begin{tabular}[c]{@{}l@{}}Time\\(ms)\end{tabular} & \begin{tabular}[c]{@{}l@{}}Ratio \\w.r.t. TL\end{tabular} & \begin{tabular}[c]{@{}l@{}}Time\\(ms)\end{tabular} & FA3 & TL \\ 
\hline
\multirow{21}{*}{\begin{sideways}\begin{tabular*}{5\normalbaselineskip}{c}Decode\end{tabular*}\end{sideways}} 
 & 8192 & 10 & 0.7 & 0.71 & 0.92 & 1.30 & 0.82 & 1.15 & 0.13 & 0.18 & 0.24 & 2.91 & 2.95 \\
 & 16384 & 10 & 1.4 & 1.39 & 1.71 & 1.23 & 1.45 & 1.04 & 0.21 & 0.15 & 0.41 & 3.40 & 3.37 \\
 & 32768 & 10 & 2.93 & 2.94 & 3.35 & 1.14 & 2.71 & 0.92 & 0.35 & 0.12 & 0.74 & 3.97 & 3.98 \\
 & 65536 & 10 & 5.85 & 5.83 & 6.74 & 1.16 & 5.39 & 0.92 & 0.65 & 0.11 & 1.43 & 4.08 & 4.07 \\
 & 131072 & 10 & 11.68 & 11.64 & 14.08 & 1.21 & 10.78 & 0.93 & 1.24 & 0.11 & 2.83 & 4.12 & 4.11 \\
 & 262144 & 10 & 21.77 & 21.63 & 25.95 & 1.20 & 20.63 & 0.95 & 2.31 & 0.11 & 5.34 & 4.08 & 4.05 \\
 & 524288 & 10 & 21.85 & 21.73 & 25.78 & 1.19 & 20.65 & 0.95 & 2.3 & 0.11 & 5.33 & 4.10 & 4.08 \\ 
\cline{2-14}
 & 8192 & 20 & 0.7 & 0.71 & 0.91 & 1.28 & 0.89 & 1.25 & 0.2 & 0.28 & 0.31 & 2.27 & 2.30 \\
 & 16384 & 20 & 1.4 & 1.39 & 1.73 & 1.24 & 1.61 & 1.16 & 0.36 & 0.26 & 0.56 & 2.50 & 2.49 \\
 & 32768 & 20 & 2.93 & 2.94 & 3.58 & 1.22 & 3.22 & 1.10 & 0.65 & 0.22 & 1.06 & 2.76 & 2.77 \\
 & 65536 & 20 & 5.85 & 5.83 & 6.97 & 1.20 & 6.22 & 1.07 & 1.24 & 0.21 & 2.04 & 2.87 & 2.86 \\
 & 131072 & 20 & 11.68 & 11.64 & 13.85 & 1.19 & 12.25 & 1.05 & 2.42 & 0.21 & 4.01 & 2.92 & 2.91 \\
 & 262144 & 20 & 21.77 & 21.63 & 27.79 & 1.28 & 24.39 & 1.13 & 4.54 & 0.21 & 7.75 & 2.81 & 2.79 \\
 & 524288 & 20 & 21.85 & 21.73 & 28.79 & 1.32 & 24.93 & 1.15 & 4.55 & 0.21 & 7.86 & 2.78 & 2.77 \\ 
\cline{2-14}
 & 8192 & 30 & 0.7 & 0.71 & 0.93 & 1.31 & 0.98 & 1.38 & 0.27 & 0.38 & 0.38 & 1.85 & 1.87 \\
 & 16384 & 30 & 1.4 & 1.39 & 1.9 & 1.37 & 1.93 & 1.39 & 0.48 & 0.35 & 0.71 & 1.98 & 1.97 \\
 & 32768 & 30 & 2.93 & 2.94 & 3.69 & 1.26 & 3.66 & 1.24 & 0.95 & 0.32 & 1.37 & 2.13 & 2.14 \\
 & 65536 & 30 & 5.85 & 5.83 & 7.19 & 1.23 & 7.06 & 1.21 & 1.82 & 0.31 & 2.64 & 2.21 & 2.21 \\
 & 131072 & 30 & 11.68 & 11.64 & 14.61 & 1.26 & 15.1 & 1.30 & 3.61 & 0.31 & 5.39 & 2.17 & 2.16 \\
 & 262144 & 30 & 21.77 & 21.63 & 28.65 & 1.32 & 27.63 & 1.28 & 6.77 & 0.31 & 10.06 & 2.16 & 2.15 \\
 & 524288 & 30 & 21.85 & 21.73 & 28.59 & 1.32 & 28.4 & 1.31 & 6.79 & 0.31 & 10.17 & 2.15 & 2.14 \\ 
\hline
\multirow{18}{*}{\begin{sideways}\begin{tabular*}{4\normalbaselineskip}{c}Prefill\end{tabular*}\end{sideways}}
 & 8192 & 10 & 0.76 & 1 & 2.01 & 2.01 & 2.01 & 2.01 & 0.36 & 0.36 & 0.62 & 1.23 & 1.62 \\
 & 16384 & 10 & 2.96 & 3.98 & 7.28 & 1.83 & 6.69 & 1.68 & 0.94 & 0.24 & 1.86 & 1.59 & 2.14 \\
 & 32768 & 10 & 12.28 & 17.13 & 28.97 & 1.69 & 25.11 & 1.47 & 2.81 & 0.16 & 6.42 & 1.91 & 2.67 \\
 & 65536 & 10 & 53.77 & 64.65 & 120.36 & 1.86 & 103.77 & 1.61 & 9.36 & 0.14 & 24.63 & 2.18 & 2.62 \\
 & 131072 & 10 & 215.76 & 262.21 & 483.69 & 1.84 & 416.53 & 1.59 & 37.18 & 0.14 & 98.55 & 2.19 & 2.66 \\
 & 262144 & 10 & 864.02 & 1048.01 & 1955.47 & 1.87 & 1696.55 & 1.62 & 160.14 & 0.15 & 408.30 & 2.12 & 2.57 \\ 
\cline{2-14}
 & 8192 & 20 & 0.76 & 1 & 2.04 & 2.04 & 2.17 & 2.17 & 0.47 & 0.47 & 0.73 & 1.04 & 1.37 \\
 & 16384 & 20 & 2.96 & 3.98 & 7.5 & 1.88 & 7.35 & 1.85 & 1.42 & 0.36 & 2.35 & 1.26 & 1.69 \\
 & 32768 & 20 & 12.28 & 17.13 & 31.25 & 1.82 & 29.78 & 1.74 & 4.68 & 0.27 & 8.65 & 1.42 & 1.98 \\
 & 65536 & 20 & 53.77 & 64.65 & 128.18 & 1.98 & 119.07 & 1.84 & 17.07 & 0.26 & 33.29 & 1.62 & 1.94 \\
 & 131072 & 20 & 215.76 & 262.21 & 507.71 & 1.94 & 476.05 & 1.82 & 72.2 & 0.28 & 136.29 & 1.58 & 1.92 \\
 & 262144 & 20 & 864.02 & 1048.01 & 2067.97 & 1.97 & 1949.78 & 1.86 & 308.36 & 0.29 & 568.53 & 1.52 & 1.84 \\ 
\cline{2-14}
 & 8192 & 30 & 0.76 & 1 & 2.12 & 2.12 & 2.36 & 2.36 & 0.59 & 0.59 & 0.86 & 0.88 & 1.16 \\
 & 16384 & 30 & 2.96 & 3.98 & 8.45 & 2.12 & 8.82 & 2.22 & 1.87 & 0.47 & 2.94 & 1.01 & 1.35 \\
 & 32768 & 30 & 12.28 & 17.13 & 32.68 & 1.91 & 33.58 & 1.96 & 6.5 & 0.38 & 10.70 & 1.15 & 1.60 \\
 & 65536 & 30 & 53.77 & 64.65 & 134.21 & 2.08 & 132.42 & 2.05 & 24.93 & 0.39 & 41.78 & 1.29 & 1.55 \\
 & 131072 & 30 & 215.76 & 262.21 & 532.39 & 2.03 & 534.61 & 2.04 & 106.12 & 0.40 & 173.00 & 1.25 & 1.52 \\
 & 262144 & 30 & 864.02 & 1048.01 & 2158.44 & 2.06 & 2192.39 & 2.09 & 457.54 & 0.44 & 727.55 & 1.19 & 1.44 \\
\bottomrule
\end{tabular}
\end{table*}

\begin{figure}[ht]
    \centering
    \includegraphics[width=\linewidth]{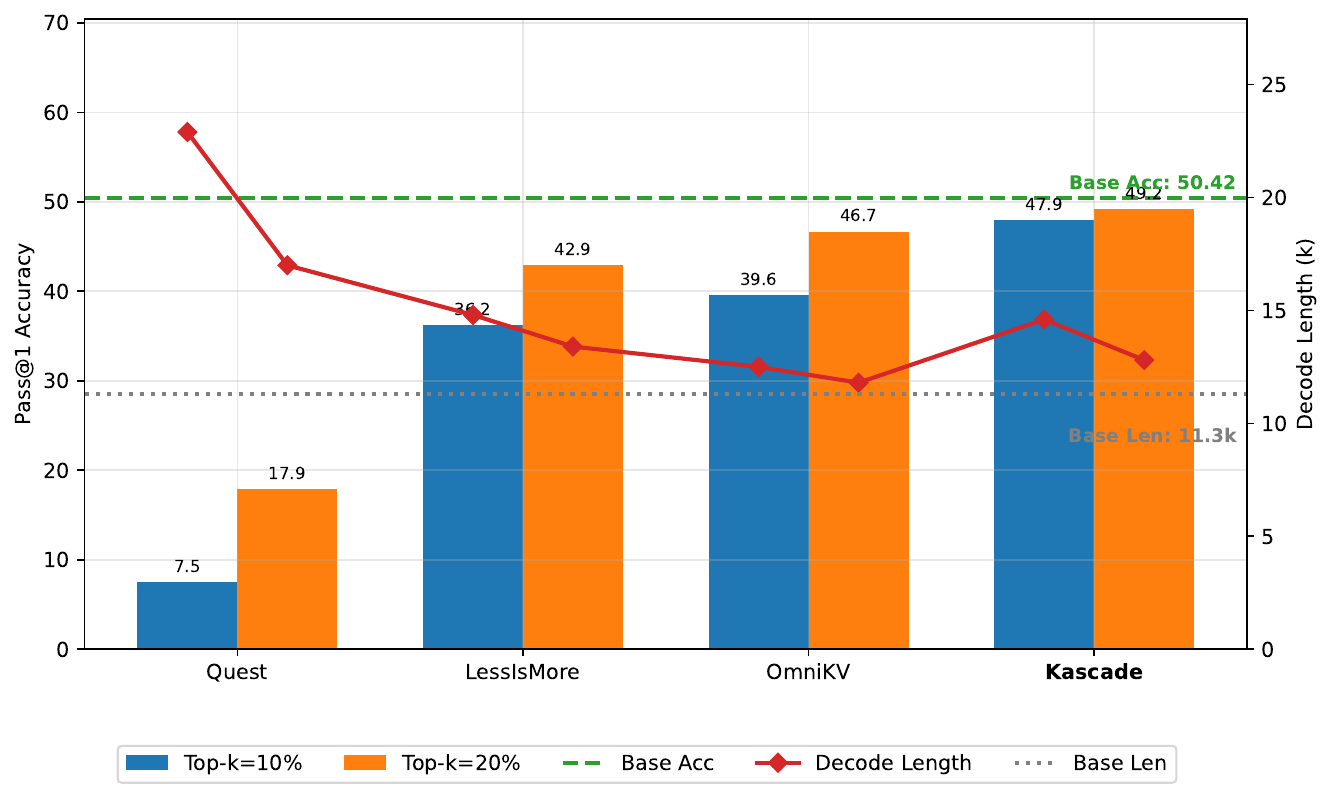}
    \caption{Accuracy and Decode lengths for Topk-$k$ 10\% and 20\%. At 20\%, \sysname is very close to the baseline and decode length also reduces significantly.  Model=\distllama, Dataset=AIME-24.}
    \label{fig:aime_vary_topk}
\end{figure}

\sysname requires choosing a set of \emph{anchor} layers. We use MuSiQue~\cite{trivedi2022musique} as a development set and the anchor layer selection algorithm in section~\ref{subsec:method_anchor_layer_selection} to choose the \emph{anchor layers}. \llama has 32 layers, of which we choose 5 anchor layers, which are [0, 2, 8, 13, 14]. \qwen has 36 layers, of which we choose 5 anchor layers - [0, 2, 7, 14, 23]. For \distllama, we use the same layers as \llama, for \sysname, OmniKV and LessIsMore. OmniKV hasn't reported the \emph{filter} layers for \qwen, so we remove it from \qwen comparisons.

For all accuracy results, we use a Top-$k$ of 10\%, with a minimum size of 128. So, if the current sequence length is $L$, during decode, the number of Top-$k$ selected is,
\[
k = min(max(0.1 \cdot L, 128), L)
\]
For \sysname, we also use the Top-$k$ in the prefill phase in a rolling manner, where each tile attends only to 10\% of previous tokens. For Quest, OmniKV, and LessIsMore, prefill phase uses full attention. For StreamingLLM, given it is a  weaker comparison, we use a sliding window size of 30\% and 4 sink tokens. For AIME-24, we also show how the accuracy changes as we increase Top-$k$ to 20\%.

\begin{figure*}[t!]
    \centering

    \begin{subfigure}{0.6\linewidth}
        \centering
        \includegraphics[width=\linewidth]{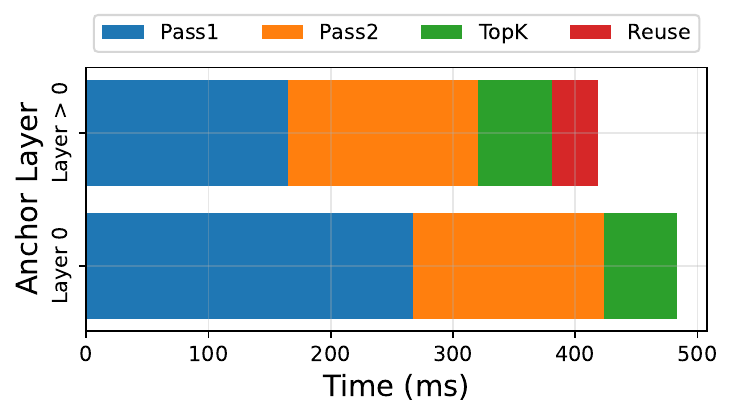}
        \caption{Prefill}
    \end{subfigure}

    %\vspace{1em}

    \begin{subfigure}{0.6\linewidth}
        \centering
        \includegraphics[width=\linewidth]{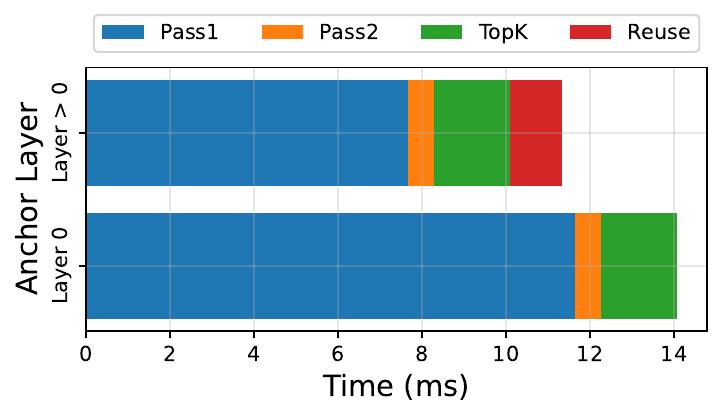}
        \caption{Decode}
    \end{subfigure}

    \caption{Time split for attention and Top-$k$ indices computation in anchor layers, in prefill and decode phase, at 128k context length for \llama setting.}
    \label{fig:anchor_layer_time_split}
\end{figure*}

\subsection{Accuracy results}

\label{subsec:eval_accuracy}

\textbf{LongBench} is composed of 21 long context tasks across 6 categories, covering multi-document QA, single-document QA, summarization, few-shot learning, code completion, and synthetic tasks. Almost all the tasks are prefill heavy, with very few decodes. All techniques except StreamingLLM and \sysname variants, do not use sparse attention in the prefill phase. Table~\ref{table:longbench} presents the results. We find that all techniques, including \sysname variants, perform very well on this benchmark. StreamingLLM is the only exception that doesn't perform well.

\textbf{AIME-24} consists of 30 challenging mathematical problems, which typically require long chain-of-thought reasoning to get to the correct answer. Models that have not been trained for reasoning perform very poorly on these tasks. Table~\ref{table:aime24} presents the average of \emph{pass@1} score, across 8 runs, for each method. The attention patterns on these tasks can be complex, and we see that StreamingLLM is unable to solve any of the problems. We find that on this complex reasoning task, \sysname performs much better than other schemes. We also show the average decode length for each method. For \sysname, the average decode length is about 29\% higher than baseline on \distllama, and 10\% higher on \qwen. We also evaluate the shared Top-$k$ across all heads variant of \sysname, and it performs worse than default \sysname across both models, but better than other schemes.

In Figure~\ref{fig:aime_vary_topk}, we evaluate the effect of increasing Top-$k$ to 20\%. \sysname continues to have the highest accuracy and is very close to the baseline. The decode length also reduces, and is about 13\% higher than baseline for \sysname.

\subsection{Efficiency results}
\label{subsec:eval_perf}

As discussed in section~\ref{subsec:method_eff}, we implemented \sysname kernels in TileLang for both prefill and decode. We ran attention microbenchmarks on a single Nvidia H100 GPU, for varying context lengths. The settings used for attention are similar to that of \llama, with 32 total number of heads, 8 key heads, 128 head dimension, in fp16. For decode benchmarks we use a batch size of 64, except at context length 512k, which uses a batch size of 32 because of insufficient memory on a single gpu. Table~\ref{table:perf_full} shows the results of these benchmarks on a combination of different Top-$k$ percentages and context lengths. We'll focus primarily on Top-$k$ percentage of 10\%, and longer context lengths. We find that our \emph{reuse} kernels see the expected speedup, and take about 10\% of the time of full attention. \emph{Anchor} layers take up more time, similar to the time of full attention. The first \emph{anchor} layer additionally does dense attention, so takes even more time. Since we have 5 \emph{anchor} layers, we compute the overall speedup accordingly. For decode phase, the speedup is about $4.1 \times$ wrt both FA3 and TileLang flashattention baselines, on \llama settings. For prefill phase, TileLang baseline kernels are about 20\% slower than FA3. Further, as discussed in section~\ref{subsec:method_eff}, our prefill kernels incur some extra costs in Top-$k$ computation, so our speedup wrt to TileLang is up to $2.6 \times$ and wrt to FA3 is up to $2.2 \times$.  For \qwen settings, since the ratio of \emph{anchor} layers is lower (5 of 36), we'd expect a higher speedup.

Figure~\ref{fig:anchor_layer_time_split} shows the time split of the multiple passes required for \emph{anchor} layers. The time for layer $0$ is higher because it does dense attention in addition to computing Top-$k$ indices. Prefill speedup takes a hit primarily because of the recomputation of attention weights in the second pass of \emph{anchor} layers.

\section{Conclusion}
We have presented \sysname as an efficient approximate Top-$k$ attention mechanism. It computes Top-$k$ indices in a subset of layers (\emph{anchor} layers), and uses them to compute Top-$k$ attention in the next few \emph{reuse} layers. To make this scheme accurate and practically deployable across models, we propose an automated way of choosing a good set of \emph{anchor} layers, and make the algorithm head-aware. We also implement efficient kernels for this scheme for both prefill and decode, which requires sharing Top-$k$ indices across a tile of tokens. \sysname is able to achieve the best accuracy on AIME-24, among other training free sparse attention schemes, at a given sparsity ratio.

There are a few limitations of this work. First, this technique requires a development set to compute the \emph{anchor} layers and head mappings. It is possible that this biases the technique towards the data in the development set. However, in the experiments we have done, we have found the selections to be robust to different datasets. Second, while \sysname reduces attention latency, it doesn't reduce the memory capacity requirements for attention. The KV caches of long sequences can be large and limit batch sizes which leads to reduced performance. Some attention works, therefore, target both capacity and latency benefits. Last,  architectures which are trained with sparsity like \cite{team2024gemma, agarwal2025gpt} will benefit less with this scheme.

\bibliography{references}
\bibliographystyle{mlsys2025}

% \appendix
% \input{sections/appendix}

\end{document}